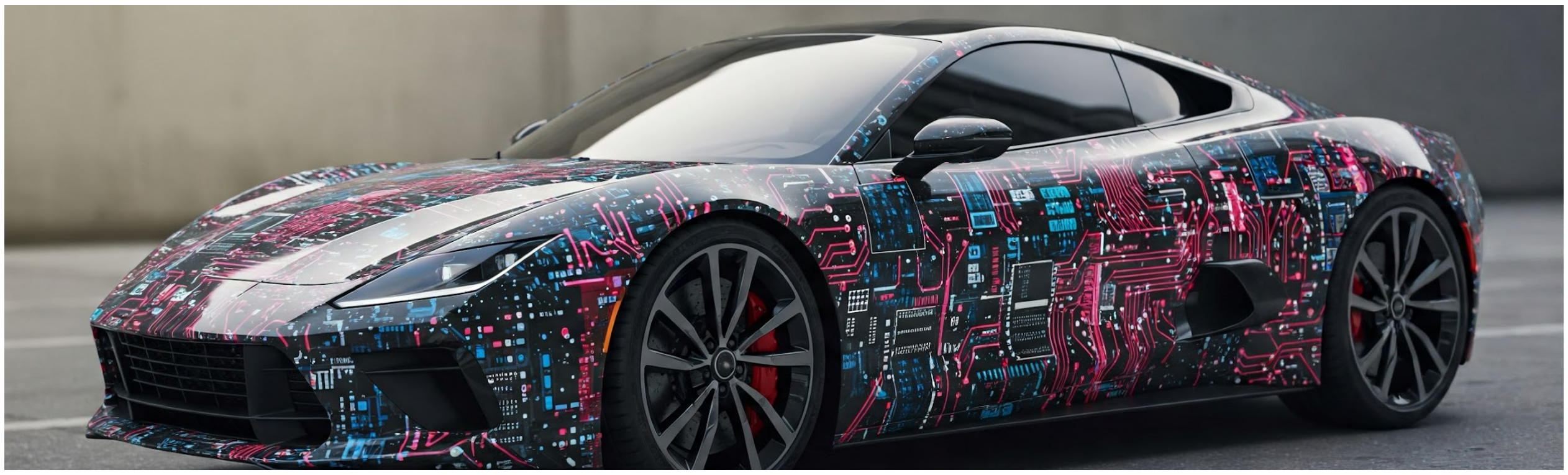

# A new wave of vehicle insurance fraud fueled by generative AI

Amir Hever, Dr. Itai Orr
UVeye Ltd.

## Abstract

Generative AI is supercharging insurance fraud by making it easier to falsify accident evidence at scale and in rapid time. Insurance fraud is a pervasive and costly problem, amounting to tens of billions of dollars in losses each year. In the vehicle insurance sector, fraud schemes have traditionally involved staged accidents, exaggerated damage, or forged documents. The rise of generative AI, including deepfake image and video generation, has introduced new methods for committing fraud at scale. Fraudsters can now fabricate highly realistic crash photos, damage evidence, and even fake identities or documents with minimal effort, exploiting AI tools to bolster false insurance claims. Insurers have begun deploying countermeasures such as AI-based deepfake detection software and enhanced verification processes to detect and mitigate these AI-driven scams. However, current mitigation strategies face significant limitations. Detection tools can suffer from false positives and negatives, and sophisticated fraudsters continuously adapt their tactics to evade automated checks. This cat-and-mouse “arms race” between generative AI and detection technology, combined with resource and cost barriers for insurers, means that combating AI-enabled insurance fraud remains an ongoing challenge. In this white paper, we present UVeye’s layered solution for vehicle fraud, representing a major leap forward in the ability to detect, mitigate and deter this new wave of fraud.

## Introduction

Insurance fraud is a massive and persistent problem, costing the United States economy hundreds of billions of dollars each year. A recent analysis estimated that over $308 billion is lost annually to insurance fraud across all lines, roughly one quarter of the industry’s total value (Vekiarides). Within the property and casualty sector (which includes auto insurance), fraud losses are about $45 billion per year, effectively adding as much as $700 in extra premiums to each American family's annual insurance costs (Hattle-Cleminshaw).

Vehicle insurance claims have long been a target for fraud through tactics like staged accidents and inflated repair bills. Now, the emergence of generative AI has dramatically expanded the scale and sophistication of this threat. Generative AI tools can produce highly realistic fake images, videos, and documents with minimal skill or effort, lowering the barrier for would-be fraudsters to manufacture convincing evidence of vehicle damage. Indeed, insurers report a surge of AI-assisted fake claims: for example, in the UK (a comparable market), one major carrier observed a 300% increase in cases of doctored auto accident photos in just a one-year period (Jones). Such "deepfake" or "shallowfake" manipulations of claim evidence are blurring the line between fact and fiction, potentially leading to enormous fraud losses if unchecked. The industry's ongoing push toward automated, "touchless" claim processing further amplifies the risk. It is projected that 70% of standard insurance claims will be handled with little or no human intervention by 2025 (Vekiarides). This efficiency gain also creates a perilous scenario as AI-manipulated images or videos could be automatically accepted by AI-driven claims systems, or bypass past traditional anti-fraud controls.

## Existing threat

One emblematic case involves the use of AI tools to forge photographic evidence of a car accident that never happened. In late 2023, investigators uncovered a fraudulent claim in which scammers had lifted a photo of a van from the owner's social media page and digitally edited it to add realistic-looking collision damage on the front bumper (Growcoot). The falsified image (an example is shown in the figure below) depicted a cracked bumper and was submitted to the insurer along with a fake repair invoice for over $1,000 in purported damages. In reality, the van had not been in any accident, the image had been seamlessly altered using generative AI-powered photo editing to create the illusion of a crash. The insurer (LV=, a UK affiliate of Allianz) grew suspicious and investigated. Tellingly, they discovered the original intact photo of the van on the policyholder's social media, identical in every way except for the added bumper cracks (Jervis). This confirmed that the claim was entirely fabricated. The attempt was thwarted, but it exemplifies how easily fraudsters can now produce photographic 'evidence' of vehicle damage out of thin air.

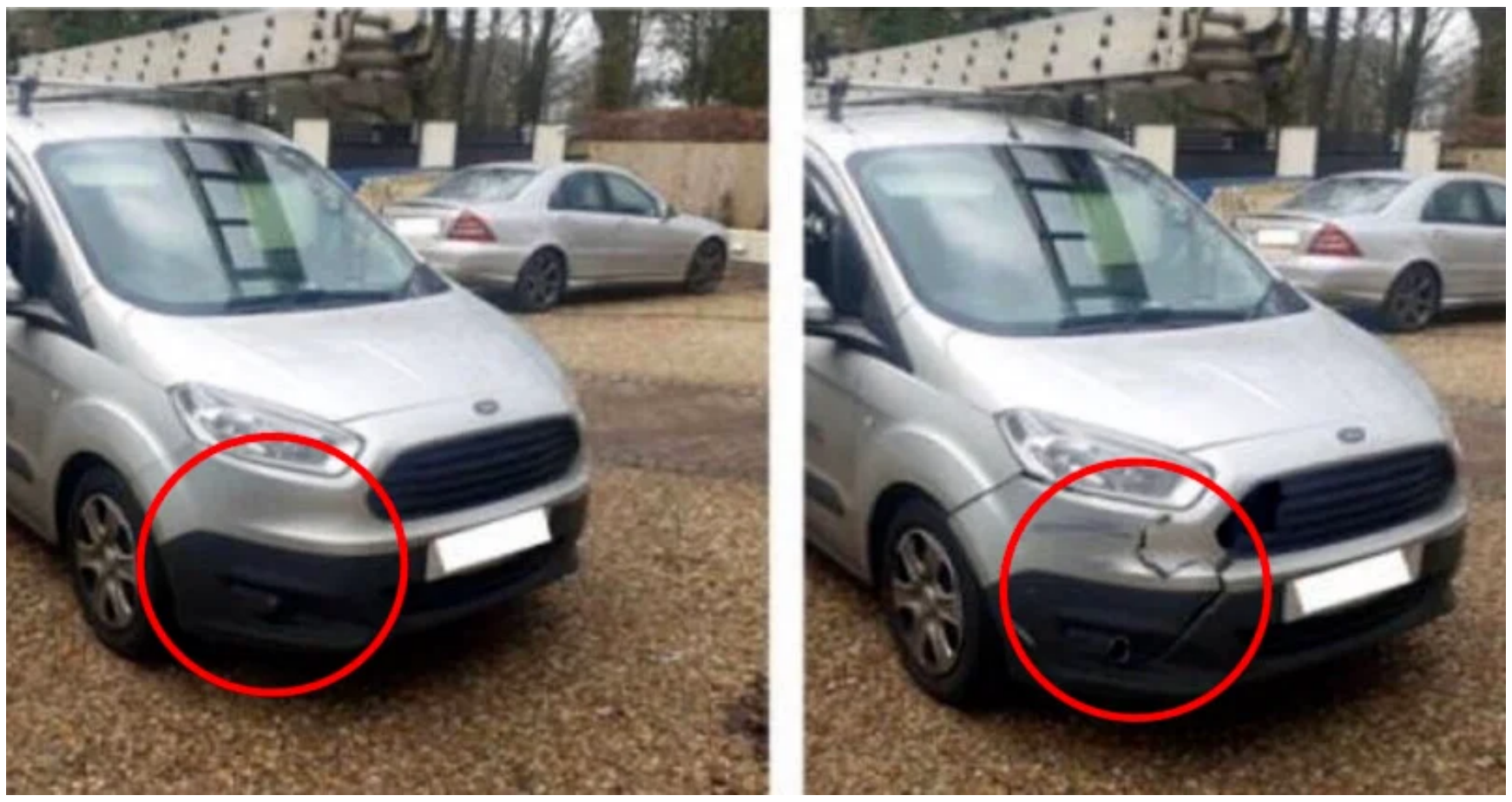

Figure 1 | Example of AI-generated claim fraud: The image on the right has been digitally altered to add a large crack in the van's bumper (highlighted) where none existed in the original photo (left). Source: (Growcoot)

Another emerging scheme involves using AI to concoct entire fictitious crash scenarios by repurposing real images of wrecked cars. In the UK, Zurich Insurance reports a trend in which fraud rings locate photographs of vehicles that were actually totaled in unrelated incidents (often found on salvage auction websites) and then use editing tools to implant a different license plate number onto the wreckage (Jones) (Jervis). The modified image makes it appear that a car belonging to the fraudster was destroyed in a crash, when in fact the vehicle in the photo is someone else's loss. Armed with these fake photos, the fraudsters file "owner" claims for total loss compensation on vehicles that were never in any accident at all. *"We have seen an increase in people locating total loss vehicles on salvage sites and then implanting a registration number onto that car. There are then claims made for that vehicle, and a claims handler would take it at face value, that it is that actual vehicle,"* explains the head of claims fraud at Zurich UK, describing this modus operandi. Such schemes effectively combine identity theft of vehicle identities with AI-assisted image manipulation, yielding entirely fabricated claims that can be difficult to detect if an adjuster simply trusts the photo and plate number. In one instance, criminals pursued a claim in an innocent person's name using a doctored image of his business van taken from the internet. Only by digging into the image's provenance did investigators reveal the deception. These cases underscore that generative technology now enables *"crash-for-cash" scams to be executed digitally*, without any real collision – fraudsters can simulate the aftermath of an accident purely through pixels and paperwork.

Recent research conducted by UVeye's research team has revealed alarming vulnerabilities in current vehicle insurance fraud detection systems. Through experimentation with readily available generative AI tools, the team successfully fabricated convincing fraudulent insurance claims, highlighting the ease with which these tools can be exploited for malicious purposes. This discovery underscores the urgent need for enhanced fraud detection mechanisms within the vehicle insurance industry to counter the evolving threat posed by sophisticated AI-driven fraud.

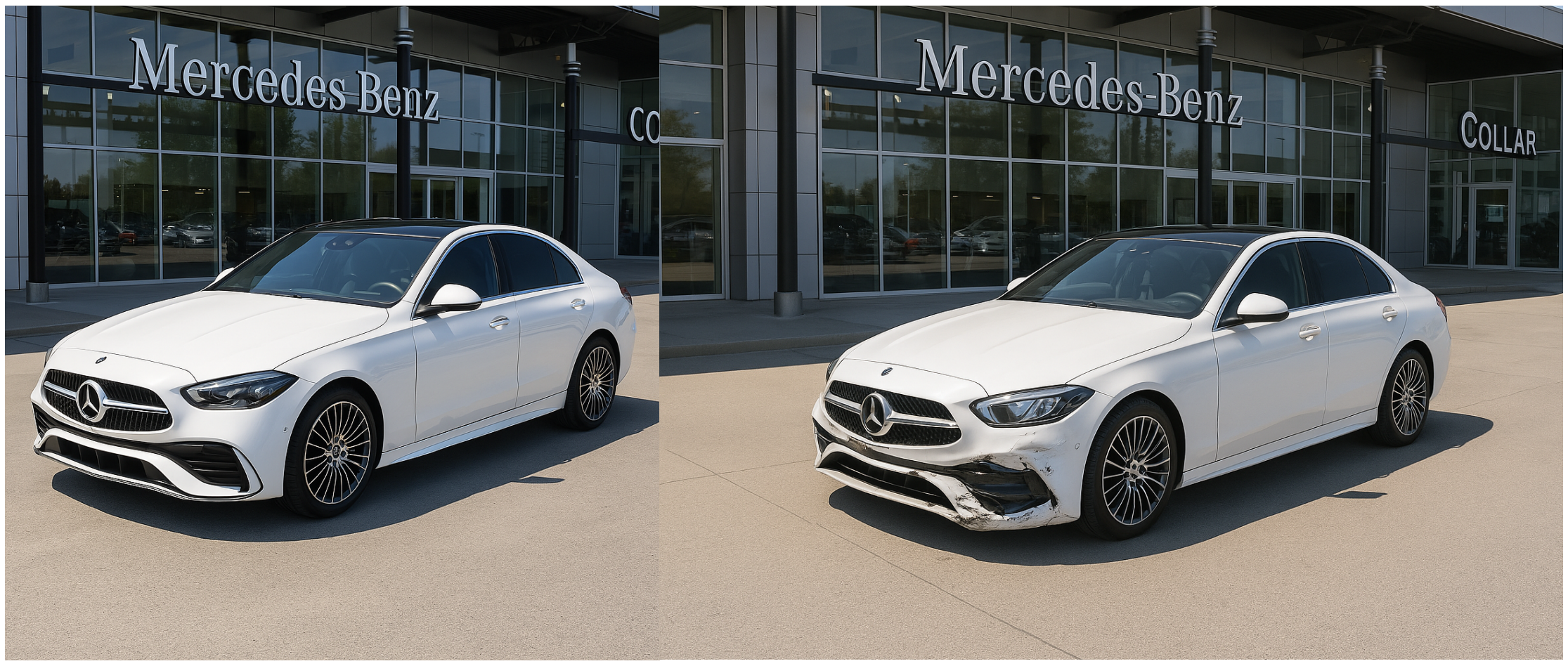

Figure 2 | Additional example of AI-generated claim fraud. On the left is the original image, on the right, an AI-generated image with damage on the vehicle's bumper where none existed in the original photo. Source: UVeye research

After generating the fake image, the researchers used a different generative AI tool to create a video from the fake image. This video depicts a walk around of the stationary car, showcasing its condition and any potential damage. The use of video further adds to the realism and potential for deception, as it can be more convincing than a static image. This combination of fake images and videos generated by AI could be used to support fraudulent insurance claims, making it appear as though damage occurred when it did not.

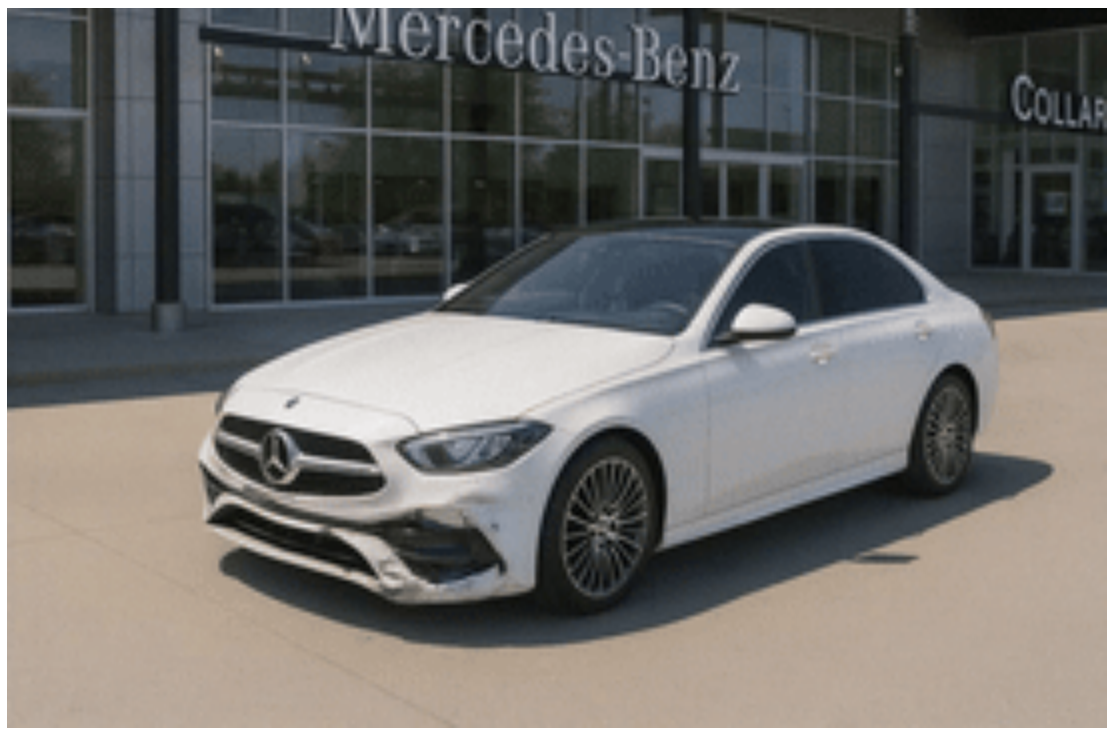

Figure 3 | The fraudulent scheme extended to the fabrication of a video depicting the purportedly damaged vehicle. This video was not a genuine recording of the incident but rather a digitally generated simulation. The seed image used to initiate and guide the video's creation was the same falsified image presented earlier in Figure 2. This manipulation technique highlights the potential for AI to not only generate static, misleading images but also to produce dynamic, seemingly authentic video evidence, further complicating the challenges of fraud detection in the insurance sector. Source: UVeye research

## Limitations of current detection and mitigation methods

Insurers are deploying a range of techniques to identify and counter generative AI-enabled vehicle insurance fraud, yet each approach has notable limitations. **AI-based image forensics** tools, including machine learning models designed to detect synthetic or manipulated images, have shown promise but remain far from foolproof. Many state-of-the-art deepfake detectors still struggle with reliability and generalization to novel forgeries (Kaur et al. #). As generative models evolve to produce more photorealistic outputs, forensic algorithms often lag behind, leading to an ongoing “arms race” between fraudsters and detectors (Kotoulas). These AI-driven detection systems are also resource-intensive and require continual updates with new training data to recognize emerging manipulation techniques, which can be costly and operationally challenging (Needham et al.). In practice, insurers find that certain AI-fabricated damage images can evade even advanced forensic checks, necessitating human expertise for final judgment.

Another traditional anti-fraud measure is **metadata analysis** of submitted photographs. Claims investigators routinely examine EXIF metadata (such as timestamps, GPS coordinates, or camera model) for inconsistencies that might signal tampering. However, this method has significant shortcomings against AI-generated content. By default, many AI-synthesized images lack the typical metadata fingerprint of an authentic camera capture (Knutsson). Even when metadata is present, it can be easily stripped or falsified by fraudsters using readily available tools, rendering it an unreliable indicator of authenticity (Pytech Academy). For example, a claimant could remove location coordinates or alter date stamps to mask the origin of an AI-generated damage photo. Because an absence or irregularity of metadata is not definitive proof of fraud – legitimate photos might lose metadata through normal processing – this technique yields at best a weak signal and can produce both false positives and false negatives.

Insurers have also introduced **workflow adjustments** to mitigate the deepfake threat, such as requiring additional documentation, manual reviews, or in-person inspections for suspicious claims. While these procedural changes can deter some fraudulent attempts, they also slow down claim processing and inflate administrative costs (Needham et al.). In an era where a majority of standard claims are now handled in a “touchless” automated fashion (Vekiarides), imposing manual checks undermines customer experience and scalability. Furthermore, relying on human adjusters to spot AI-crafted forgeries is increasingly difficult as the fakes become nearly indistinguishable from real evidence (Kotoulas). Even highly trained experts can be deceived by high-quality synthetic images, so purely manual safeguards provide imperfect protection; moreover, insurers cannot simply add unlimited verification steps without unduly burdening legitimate claimants.

Finally, **industry collaboration** has emerged as an important strategy, with insurers sharing information on suspected fraud cases and collectively developing countermeasures. Initiatives such as cross-company fraud databases and image-sharing repositories can help identify repeat offenders and detect scam patterns across insurers. However, this collaborative approach faces its own hurdles. Data sharing between companies remains limited, siloed information systems and privacy constraints impede a seamless exchange of fraud intelligence (synectics solutions). Not all carriers are equally willing or able to contribute data, due to competitive sensitivities and legal restrictions on customer information. As a result, fraudsters may exploit gaps between organizations, for instance, recycling the same AI-fabricated claim at multiple insurers that do not communicate in real time. The lack of standardization and real-time coordination means the industry's defense against generative AI fraud is only as strong as its weakest link. While significant efforts are underway to adapt to AI-enabled insurance fraud, current detection and mitigation methods each have inherent limitations that leave insurers struggling to keep pace with increasingly sophisticated fraudulent techniques.

## Layered Security Solution powered by UVeye

Taking lessons from the limitations mentioned above, UVeye's comprehensive three-layer Security Solution is designed to combat vehicle claim fraud, providing a trusted and reliable method for verifying claims and detecting fraudulent activity.

- The first layer of security involves the use of a **trusted third party** to oversee the claims process. This ensures that all claims are handled fairly and impartially, reducing the risk of bias or collusion.
- The second layer utilizes an **automated scanning system** that incorporates multiple cameras and multiple frames to capture detailed images of the vehicle. This allows for a thorough inspection of the vehicle's condition, identifying any damage or inconsistencies that may indicate fraud.
- The third layer involves the **embedding of an encrypted digital fingerprint** within the vehicle's data. This fingerprint serves as a tamper-proof record of the vehicle's condition, making it impossible for fraudsters to alter or manipulate the data without detection.

By combining these three layers of security, UVeye provides a robust and comprehensive solution for detecting and preventing vehicle claim fraud. This innovative approach not only protects insurance companies from financial loss but also helps to maintain the integrity of the insurance industry as a whole.

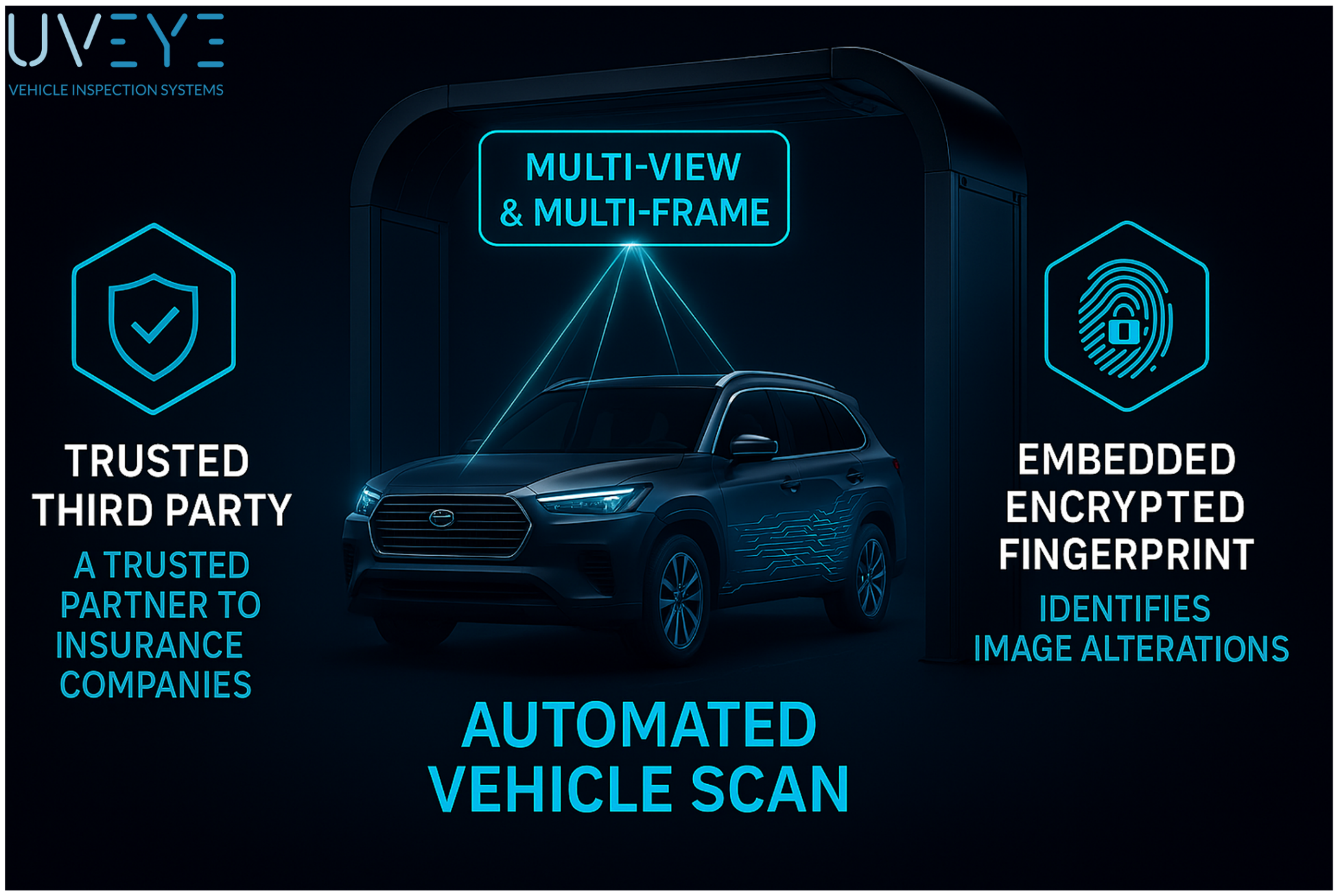

Figure 4 | UVeye's three-layer Security Solution for vehicle claim fraud. Trusted third party, automated scan consists of multi-camera and multi-frame and embedded encrypted digital fingerprint to detect tampering. Source: UVeye

Shown in Figure 5, the integration with an automated claim system modifies the flow and adds an automated scan step as verification.

1. **Step 1: Claim submission:** The claim process is initiated when a policyholder submits a claim for vehicle damage, typically through an online portal, mobile app, or by contacting their insurance provider. The claim includes details about the incident, the extent of the damage, and any supporting documentation, such as photos or police reports.
2. **Step 2: Automated vehicle scan:** Once the claim is submitted, a policyholder is instructed to scan the vehicle in the nearest UVeye scanner location.
3. **Step 3: AI-powered damage assessment:** The images captured during the scan are analyzed by an AI-powered system. This system is trained on a vast dataset of vehicle images and damage assessments, allowing it to accurately identify and classify different types of damage, such as dents, scratches, and broken parts. The AI also estimates the severity of the damage and calculates an approximate repair cost.
4. **Step 4: Claim verification:** The AI's damage assessment is then compared to the information provided by the policyholder in their claim. If the AI's findings match the reported damage, the claim is considered verified and proceeds to the next stage of the claims process.
5. **Step 5: Fraud detection:** If the AI's assessment does not match the reported damage, the system flags the claim as potential fraud. This could indicate that the policyholder has exaggerated the damage, submitted photos of pre-existing damage, or even attempted to stage an accident. The flagged claim is then routed to a claim's investigator for further review.
6. **Step 6: Claim processing:** Verified claims are processed according to the insurance company's standard procedures. This typically involves issuing a payment to the policyholder or their chosen repair shop, minus any applicable deductibles.

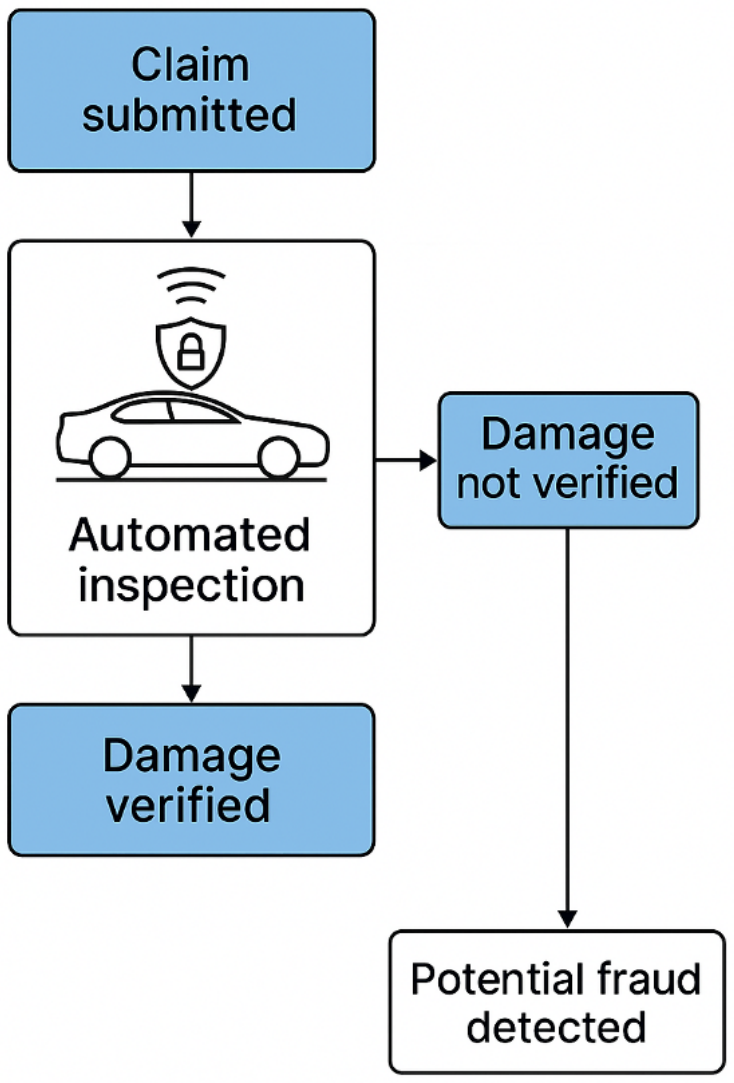

Figure 5 | Automated claim verification flow using AI-powered vehicle inspection. After a claim is submitted, an automated scan of the vehicle authenticates the reported damage. If damage is verified, the claim proceeds; if not, the system flags the case as potential fraud, ensuring claims are based on physical evidence rather than potentially manipulated images. Source: UVeye

By utilizing UVeye's fraud suppression solution, insurers gain access to multiple advantages:

1. **Verification of claimed damage:** UVeye's physical scans offer insurers an objective benchmark, immediately verifying whether claimed damages exist. Deepfake or edited claimant photos cannot alter the real-world state of the vehicle. This prevents payouts on fabricated claims and discourages fraud attempts relying solely on manipulated media
2. **Baseline vehicle records:** Pre-policy or pre-claim scans create a verifiable condition history, allowing insurers to identify pre-existing damage and prevent customers from falsely attributing old issues to new incidents. Automated image comparisons between pre- and post-incident scans highlight discrepancies clearly.
3. **Post-accident damage assessment:** Following an incident, a UVeye scan delivers a comprehensive and unbiased damage report, ensuring claim accuracy and reducing inflated estimates from repairers or claimants. UVeye's AI also maps damage severity to streamline claim settlements.
4. **Detection of staged or exaggerated claims:** UVeye uncovers physical inconsistencies that reveal staged events, such as undercarriage scraping that contradicts an accident narrative or damage patterns inconsistent with the described collision. These insights help differentiate genuine claims from orchestrated fraud.
5. **Reducing reliance on claimant evidence:** By incorporating UVeye scans into claims workflows, insurers reduce dependency on customer-submitted photos, which are increasingly unreliable due to deepfakes and image manipulation. UVeye provides trusted, tamper-proof evidence, enhancing claims integrity.
6. **Scalable integration:** UVeye systems are already deployed at dealerships and inspection centers and can be integrated at preferred repair networks or mobile units for field use. Secured cloud-based data sharing allows insurers to instantly access inspection results, cross-reference historical scans, and embed findings directly into claims management systems.

## UVeye Case Study: Acadia Insurance Pilot Program

UVeye has partnered with Acadia Insurance, a W. R. Berkley Company, to pilot a next-generation vehicle inspection process aimed at accelerating claims handling while enhancing accuracy and fraud prevention. The initiative integrates UVeye's three-layer scanning system into Acadia's workflow, generating meaningful operational improvements.

At participating dealerships, customers simply drive through a UVeye system. The entire scan takes just seconds, eliminating the need to upload photos or wait for a manual appraisal. The system captures a high-resolution, multi-angle image set, including underbody views, which is automatically transmitted to Acadia's appraisers. Because the scans are performed by a trusted third party and cryptographically signed, the process offers strong protection against tampering or selective photo submissions.

The detailed image capture provides visibility that appraisers often lack in traditional field inspections. As Barbara Taylor, Director of Auto Physical Damage Claims at Acadia, explains:

> *"Our appraisers remark that they would be unlikely to see the damage in such detail in the field, particularly underneath the vehicle, which often would result in the need for re-inspection and a supplemental estimate once the vehicle was in a repair shop and up on a lift."*

The pilot has already demonstrated key benefits: cycle times have shortened due to the elimination of supplemental inspections, prior or unrelated damage is being identified earlier in the process, and estimated accuracy has matched or exceeded that of conventional field appraisals.

Looking ahead, Acadia and UVeye are exploring the use of AI to generate real-time, fully cost estimates based on scan data. Taylor notes the potential impact:

> *"Think of the gains in turn-around time if a customer can simply drive through a UVeye system and have a completed appraisal ready almost instantaneously! This could be a game-changer for auto-claim cycle time, and in turn, a huge boost for customer service."*

By embedding UVeye's automated scans directly into the claims process, Acadia is demonstrating how insurers can both reduce fraud risk and expedite legitimate claims, delivering a faster, more transparent experience for policyholders and carriers alike.

## Industry collaboration

UVeye's layered solution is a leading example of how industry collaboration can address systemic challenges in insurance, particularly around fraud prevention, data integrity, and operational efficiency. By integrating with initiatives like RAPID X, which enables secure, permissioned exchange of first-notice-of-loss data among carriers during mutual events, UVeye strengthens the foundation for faster, more accurate, and more secure claims handling. Its trusted, cryptographically signed scan data helps carriers make confident, evidence-based decisions while minimizing the risk of manipulation.

Additionally, collaboration with RiskStream's AI Council, a cross-industry working group of insurers, insurtechs, and research institutions can also strengthen the overall solution. Together, these initiatives can identify high-impact AI use cases, such as automated damage detection and fraud prevention, and promote ethical standards for data use.

As generative AI models continue to evolve, the insurance sector will require infrastructure-level authenticity verification, not just content-level detection. UVeye's scanning and digital fingerprinting model exemplifies this shift toward physical-grounded AI integrity.

## Conclusions

Generative AI has rapidly accelerated the sophistication and scale of vehicle insurance fraud. Fraudsters now easily fabricate damaged photos, videos, and documents, undermining traditional claim verification processes. While insurers have deployed AI detection tools, metadata checks, and manual reviews, each method faces limitations: evolving AI models evade forensic detection, metadata can be stripped or falsified, and human reviews are costly and error prone. Industry collaboration helps but remains fragmented, leaving critical gaps.

UVeye's three-layer solution offers a transformative defense. By anchoring claims to physical, multi-frame, multi-camera scans, insurers gain objective, tamper-proof evidence of vehicle condition. The embedded encrypted fingerprint in scan data further secures authenticity, detecting any attempt at post-scan tampering. As a trusted third party, UVeye adds independent verification to insurer workflows, reducing reliance on claimant-provided photos or documents susceptible to manipulation.

Beyond detection, UVeye acts as a deterrent. Knowing claims will face automated, impartial inspection discourages fraud attempts from the outset, while genuine claims benefit from faster, more reliable resolution. Integrated at key points, policy issuance, post-accident assessment, repair validation, UVeye strengthens fraud defenses while improving customer experience.

By embedding objective physical verification into the heart of claims processing, UVeye offers insurers a scalable path to reduce fraud, lower costs, and restore trust in the claims process, turning the tide against the growing threat of generative AI.